\documentclass{article}
\usepackage{iclr2027_conference,times}

\usepackage{amsmath,amsfonts,bm}

\def\eqref#1{equation~\ref{#1}}

\def\1{\bm{1}}

\DeclareMathAlphabet{\mathsfit}{\encodingdefault}{\sfdefault}{m}{sl}
\SetMathAlphabet{\mathsfit}{bold}{\encodingdefault}{\sfdefault}{bx}{n}

\usepackage{hyperref}
\hypersetup{
  hypertexnames=false,
  hidelinks,
  pdftitle={Steer, Don't Tell: Breaking Reasoning Plateaus with Reachability-Aligned Self-Distillation},
  pdfauthor={Zijian Wu, Lingkai Kong, Lin Yang, Haiteng Zhao, Kuikun Liu, Kai Chen, Dahua Lin, Wenwei Zhang}
}
\usepackage{url}
\usepackage{amsmath,amssymb,amsfonts}
\usepackage{amsthm}
\usepackage{algorithm}
\usepackage{algorithmic}
\usepackage{graphicx}
\usepackage{booktabs}
\usepackage{multirow}
\usepackage{xcolor}
\usepackage{colortbl}
\usepackage{xfp}
\usepackage{capt-of}
\usepackage{fvextra}
\DefineVerbatimEnvironment{Prompt}{Verbatim}{
  fontsize=\small,
  breaklines=true,
  breakanywhere=false,
  breakindent=0pt,
  breakautoindent=false,
  breaksymbolleft={},
  breaksymbolright={},
  frame=single,
  framerule=0.35pt,
  framesep=3pt,
  rulecolor=\color{black!35},
  formatcom=\color{black!80}
}

\definecolor{ourstint}{HTML}{E7F0FB}
\definecolor{deltagreen}{HTML}{1B7A3D}
\definecolor{deltared}{HTML}{B3123F}

\newcommand{\oursrow}{\rowcolor{ourstint}}
\newcommand{\dlt}[1]{\textcolor{deltagreen}{\small $+#1$}}

\newcommand{\dbad}[1]{\textcolor{deltared}{\small $#1$}}
\newcommand{\best}[1]{\textbf{#1}}
\newcommand{\snd}[1]{\underline{#1}}

\definecolor{barblue}{HTML}{2A78D6}
\definecolor{barref}{HTML}{8C8B87}
\definecolor{barpale}{HTML}{DEDDD9}
\definecolor{tabband}{HTML}{F4F3F0}
\definecolor{tabfaint}{HTML}{8C8B87}

\newcommand{\barlo}{0}
\newcommand{\barhi}{100}
\newcommand{\barmm}{12}
\newcommand{\barnw}{2.6em}
\newcommand{\barscale}[4]{\renewcommand{\barlo}{#1}\renewcommand{\barhi}{#2}%
  \renewcommand{\barmm}{#3}\renewcommand{\barnw}{#4}}
\newcommand{\vbar}[2]{\textcolor{#2}{\rule[0.08ex]%
  {\fpeval{max(0.35, min(1, (#1 - \barlo) / (\barhi - \barlo)) * \barmm)}mm}%
  {0.58ex}}}
\newcommand{\barcell}[3]{\makebox[\barnw][r]{#1}\hspace{2.6pt}%
  \makebox[\barmm mm][l]{\vbar{#2}{#3}}}
\newcommand{\barours}[1]{\barcell{\textbf{#1}}{#1}{barblue}}
\newcommand{\barrefr}[1]{\barcell{#1}{#1}{barref}}
\newcommand{\barbase}[1]{\barcell{#1}{#1}{barpale}}
\newcommand{\barsnd}[1]{\barcell{\underline{#1}}{#1}{barpale}}

\newcommand{\blockhead}[2]{\rowcolor{tabband}\multicolumn{#1}{@{}l}{\textbf{#2}}}

\newcommand{\SSR}{\textsc{SSR}}
\newcommand{\methodname}{Self-Steered Reasoner}

\title{Steer, Don't Tell: Breaking Reasoning\\
Plateaus with Reachability-Aligned\\
Self-Distillation}

\author{
  \textbf{Zijian Wu}$^{1,2,*}$ \quad
  \textbf{Lingkai Kong}$^{2,3,*}$ \quad
  \textbf{Lin Yang}$^{2}$ \quad
  \textbf{Haiteng Zhao}$^{2}$ \\
  \textbf{Kuikun Liu}$^{2}$ \quad
  \textbf{Kai Chen}$^{2}$ \quad
  \textbf{Dahua Lin}$^{1,2,\dagger}$ \quad
  \textbf{Wenwei Zhang}$^{2,\dagger}$ \\[0.7em]
  {\normalfont\small $^{1}$ Department of Information Engineering, The Chinese University of Hong Kong} \\
  {\normalfont\small $^{2}$ Shanghai Artificial Intelligence Laboratory} \\
  {\normalfont\small $^{3}$ Shanghai Jiao Tong University} \\[0.35em]
  {\normalfont\small $^{*}$ Equal contribution. \quad $^{\dagger}$ Corresponding authors.} \\
  {\normalfont\small \texttt{dhlin@ie.cuhk.edu.hk} \quad \texttt{wayne.zw@outlook.com}}
}

\begin{document}

\maketitle

\begin{abstract}
Post training for large reasoning models remains insufficient on challenging problems, whose correct trajectories are hard to sample. Though prior works address this difficulty by integrating supervised fine tuning on expert trajectories with reinforcement learning, such external experts are not always available. Existing self improvement methods instead use reference solutions to create a privileged self-teacher. Yet privileged teacher correctness does not imply student reachability. Solution conditioning shifts the teacher distribution toward actions unlikely under the unprivileged student, producing seemingly "correct" trajectories that are unreachable or even mal-formed learning targets. Accordingly, we identify reachability alignment as the key to self constructed supervision and propose \methodname{} (\SSR{}). 
SSR realizes this principle through student support projection, which transforms the privileged teacher distribution into a directional signal only over reasoning actions locally reachable under the unguided student. Verification then retains correct trajectories sampled from this reachable space. To further boost the sampling efficiency, an iterative refinement framework is adopted, using diagnostic guidance based on current trajectories. 
On problems with pass@64 = 0 even after extensive RL, SSR recovers correct and learnable trajectories for 74.6\% without parameter updates. Fine tuning on these trajectories followed by resumed reinforcement learning improves aggregate accuracy by 4.9 percentage points on AIME25/26 and HMMT26. These results show that privileged guidance becomes effective self-improvement data when aligned with the reasoning actions the student can reach.
\end{abstract}

\section{Introduction}
Training large reasoning models on verifiable tasks has driven substantial gains in mathematical and scientific problem solving~\citep{openai2024o1,deepseekr1,shao2025deepseekmathv2}. Yet reinforcement learning remains ineffective on the hardest problems when correct trajectories are rarely sampled~\citep{chen2026nurl,li2025questa,chen2026unlearnability}. Prior work show that supervised or off-policy expert guidance can complement
reinforcement learning on such questions
~\citep{ma2025relift,zhang2025chord,yan2025luffy}. Their supervision, however,
is supplied by stronger models or external expert trajectories, which may be
unavailable when training frontier reasoning models. This motivates a self contained setting in which the model must improve using only the problem and its reference solution.

Prior self improvement methods instead use reference solutions to construct supervision. Offline methods condition generation on solutions to synthesize rationales~\citep{zelikman2022star,dail2026,hinteddecoding2026,dasd2026}, while online privileged self distillation methods construct a teacher with access to complete solutions for token level supervision~\citep{zhao2026opsd,jiang2026trd}. In either case, reference conditioning gives the same model the intended outcome, making trajectories that appears to have a correct outcome easier to produce. Yet privileged teacher's outcome correctness does not imply student reachability. The teacher may exploit the final answer and intermediate results to construct reasoning actions that are unlikely or even illegal(e.g. referencing a non-existing hint) under the unprivileged student. Its trajectory can therefore be correct under privileged conditioning while remaining unreachable or even malformed as a learning target. Directly imitating the teacher will transfers such behavior to the student. Explicit answer leakage,
shortcut hacking, and post hoc rationalization are visible consequences therefore.

\begin{figure}[t]
\centering
\includegraphics[width=\textwidth]{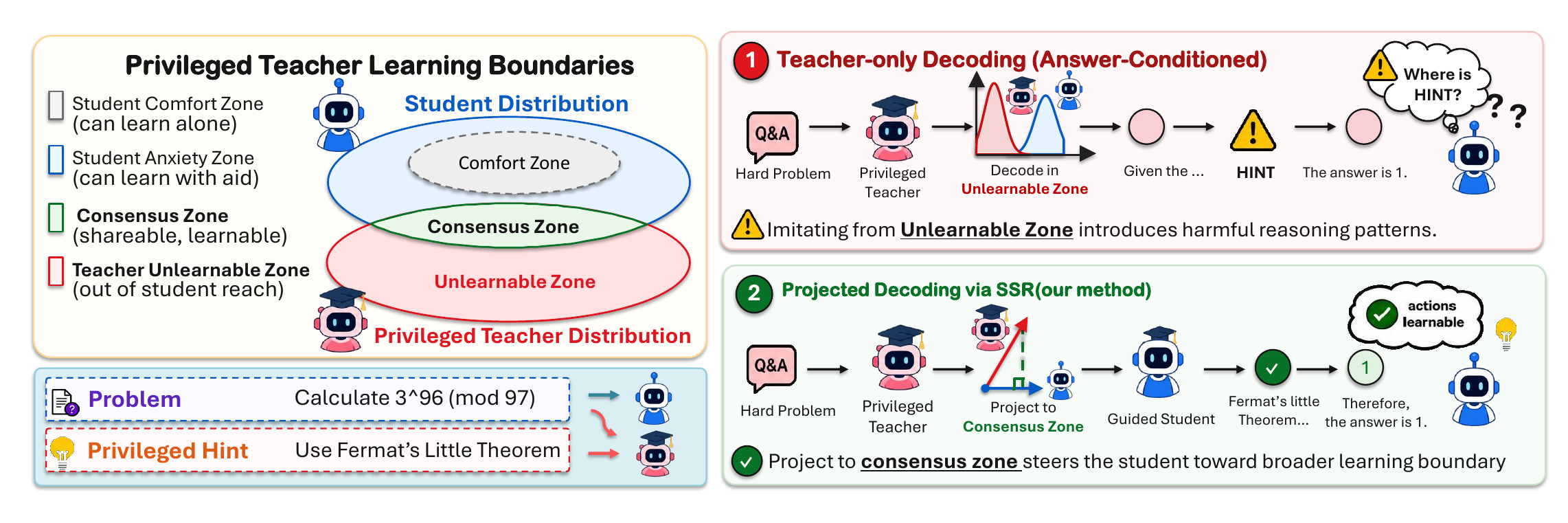}
\caption{
RLVR can extend the model's ability in its comfort zone, but it meets obstacles when coming to anxiety zone where correct trajectories are hard to sample. Privileged-teacher can provide aid signals on anxiety zone, but it can also select actions outside the student's reachable region, producing unlearnable traces. Directly imitating these actions can corrupt the student's reasoning ability(e.g., learning to hallucinate a non-existent hint). By projecting privileged preferences into the consensus region, \SSR{} retains the learnable steering signals within the privileged preferences, effectively expands the student's learning boundary.
}
\label{fig:teaser}
\end{figure}

We identify this correctness reachability mismatch as the central obstacle to self improvement and propose \methodname{} (\SSR{}). The core principle is to align the privileged teacher with what the student can reach. At each decoding prefix,
student-support projection restricts the guided teacher to actions that remain
plausible under the unguided student, while preserving the teacher's relative
preferences among those actions. The student defines the reachable action set,
and the teacher supplies a direction within it. The joint-decoding mechanism drives the student into a more plausible trajectory. Verification subsequently
retains correct completed trajectories. This separation turns a privileged
solution from an imitation target into guidance for searching the student's
reachable reasoning space.

To further boost the search efficiency, \SSR{} places projected decoding inside an iterative refinement framework. Each round
constructs a diagnostic steer from the current failed trajectory and the
reference solution.
The steer records the error and an abstract corrective
direction while excluding the final answer and unseen numerical values.
Guided and unguided next-token distributions are then compared along the
failed trajectory, and their KL disagreement selects an informative
intervention frontier. \SSR{} retains the preceding prefix and regenerates
only the suffix through projection. An incorrect continuation enters the next
round with a new diagnosis and intervention point.

Our experiments show that student support projection effectively reduces hallucination on privileged information and general hindsight bias.
The complete \SSR{} framework recovers verified correct trajectories for 147 of the 197 problems where the plateaued Qwen3-8B model achieves \texttt{pass@64}$=0$, or 74.6\%, without parameter updates. Fine-tuning on the resulting trajectories followed by
resumed RL improves aggregate accuracy by 4.9 percentage points on Qwen3-8B
and 5.2 points on Qwen3-4B-Instruct. Ablation experiments further shows that student-support projection
substantially reduces hindsight anchoring effect and hint leakage.
Our contributions are as follows.
\begin{itemize}
    \item We identify the correctness reachability mismatch in privileged self distillation and formulate reachability alignment as the key requirement for self constructed reasoning supervision.

    \item We introduce student-support projection and instantiate it in
    \SSR{} through diagnostic steering, divergence-guided intervention,
    projected continuation, verification, and iterative refinement.

    \item We isolate the effect of projection under fixed guidance and show
    that the complete framework recovers verified trajectories for 74.6\% of
    the empirically unsolved set, producing supervision that enables resumed
    RL to move beyond the parent plateau.
\end{itemize}

\section{Method}
\label{sec:method}

We introduce \methodname{} (\SSR{}), a self distillation framework for constructing supervision on problems where a plateaued reasoning policy receives no positive RL signal. The core operation is student support projection, which restricts a privileged teacher to reasoning actions locally reachable under the unguided student. \SSR{} places this operator inside an iterative refinement procedure that diagnoses the current failure, selects an informative intervention point, and regenerates only the remaining suffix.

\begin{figure}[t]
\centering
\includegraphics[width=\textwidth]{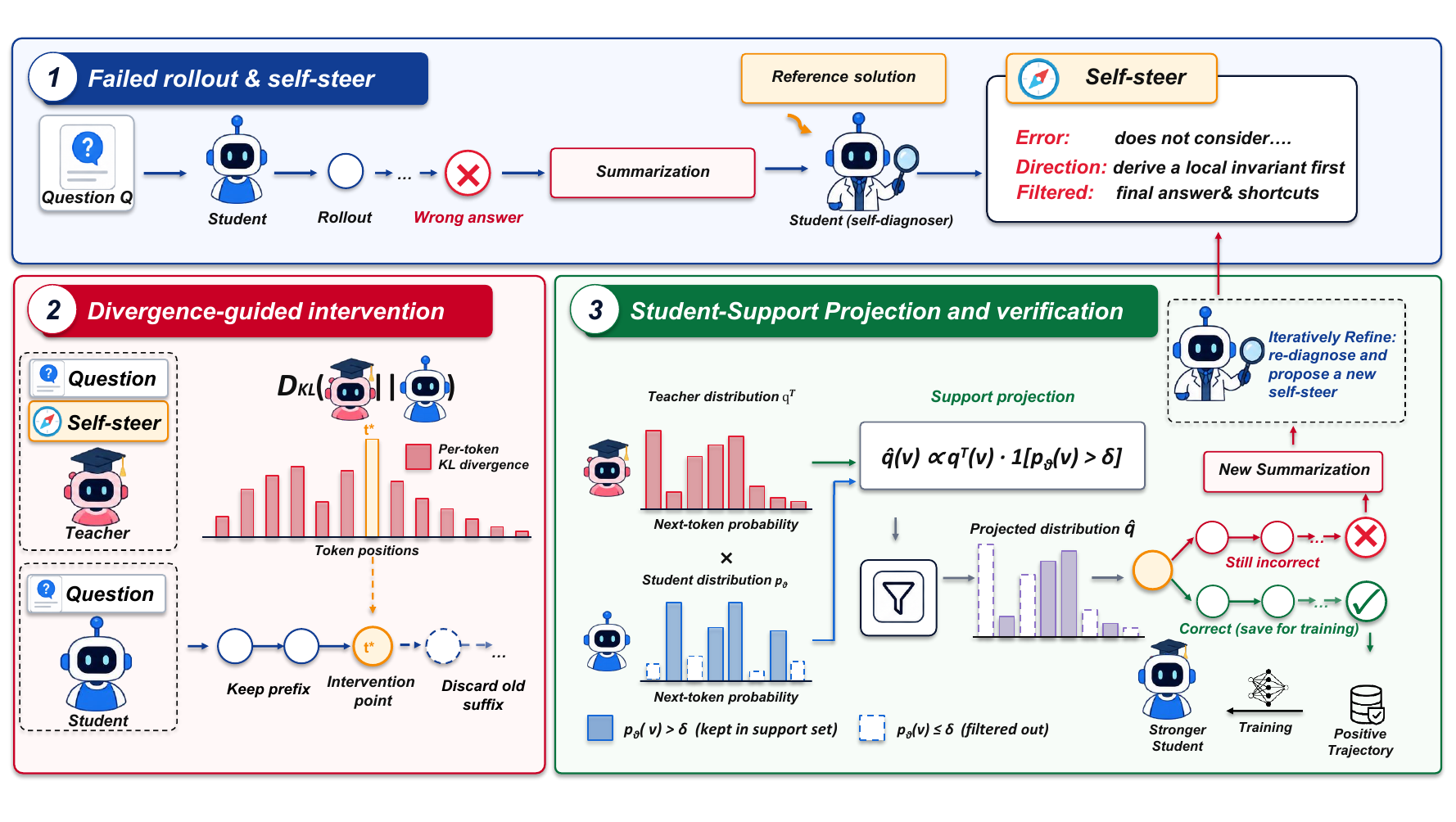}
\caption{\textbf{Overview of the \SSR{} framework.}
1. Given a very hard question Q, we sample failed rollouts from the student model.
The failed rollout is then summarized and diagnosed against the reference solution
to produce a self-steer. The self-steer describes the error and provide fixing directions while avoiding directly offering the answer. 2. The self-steer is then used to construct the privileged self-teacher. The KL-divergence between teacher and student is calculated on the failed rollout to determine the proper intervention point. Everything after that point is discarded. 3. After the intervention point, for every decoding prefix, the student first calculate its support-set by retaining every token with a student possibility above a certain threshold $\delta$. The teacher then rewights these tokens and choose the next token. For each new trajectory generated, it will be stored for further training if correct, and send back to a new round of refinement if incorrect. After the iterative refinement succeeds, the saved positive trajectory can be used to improve the student model.}
\label{fig:ssr_pipeline}
\end{figure}

\subsection{Post-Plateau Self-Improvement}
\label{sec:problem_setting}

Let $p_{\theta_0}$ denote a reasoning policy after RL training has reached a
performance plateau. Each verifiable problem $x$ is paired with a reference
solution $g_x$, and $R(x,y)\in\{0,1\}$ evaluates whether a generated
trajectory $y$ is correct. We define the empirically unsolved set under a
rollout budget $K$ as
\begin{equation}
\mathcal{H}_{\theta_0}^{K}
=
\left\{
x
\;\middle|\;
\max_{1\leq k\leq K}
R(x,y_k)=0,
\quad
y_k\sim p_{\theta_0}(\cdot\mid x)
\right\}.
\label{eq:hard_set}
\end{equation}
These problems provide no positive reward under the specified sampling
budget. Our goal is to construct verified trajectories for them using the
frozen policy and their reference solutions.
Conditioning the policy on $g_x$ creates a privileged self-teacher that can
produce correct trajectories more readily. The additional context can also
shift probability toward actions that are locally implausible under the
unprivileged policy. We seek trajectories that combine verified correctness
with local student reachability.

\subsection{Student-Support Projection}
\label{sec:support_projection}

Let $s$ denote the problem and the reasoning prefix generated so far. We
define the student's local action set using a probability floor $\delta$
\begin{equation}
\mathcal{V}_{\theta_0,\delta}(s)
=
\left\{
v
\;\middle|\;
p_{\theta_0}(v\mid s)>\delta
\right\}.
\label{eq:student_support}
\end{equation}
Throughout the paper, student support refers to this thresholded local set.

Let $q_d(v\mid s)$ denote the next-token distribution of the same frozen
model when additionally conditioned on a filtered diagnostic steer $d$.
Direct sampling from $q_d$ retains its corrective preferences together with
actions that may lie outside the student's local support. We define
student-support projection as
\begin{equation}
\Pi_{\theta_0,\delta}[q_d](v\mid s)
=
\frac{
q_d(v\mid s)
\mathbf{1}
\left[
p_{\theta_0}(v\mid s)>\delta
\right]
}{
\sum_u
q_d(u\mid s)
\mathbf{1}
\left[
p_{\theta_0}(u\mid s)>\delta
\right]
}.
\label{eq:support_projection}
\end{equation}
The student determines the admissible action set, while the guided teacher
preserves its relative preferences among the remaining actions. We apply
~\eqref{eq:support_projection} autoregressively and use $\delta=0.02$ in
all experiments. The choice of this hyperparameter is presented in the appendix.

\paragraph{Projection interpretation.}
For any prefix where the denominator in
~\eqref{eq:support_projection} is nonzero, the projected distribution is
the KL projection of $q_d$ onto distributions supported by
$\mathcal{V}_{\theta_0,\delta}(s)$
\begin{equation}
\Pi_{\theta_0,\delta}[q_d](\cdot\mid s)
=
\arg\min_{
r:\operatorname{supp}(r)
\subseteq
\mathcal{V}_{\theta_0,\delta}(s)
}
D_{\mathrm{KL}}
\left(
r(\cdot\mid s)
\middle\|
q_d(\cdot\mid s)
\right).
\label{eq:projection_property}
\end{equation}
The proof is given in Appendix~\ref{app:projection_proof}. Projection
enforces local reachability, and the verifier evaluates sequence-level
correctness.

\subsection{Iterative Projected Refinement}
\label{sec:iterative_refinement}

A projected decoder still needs a correction direction and an intervention
point. One correction can also expose an error that was hidden by the
original failure. \SSR{} therefore applies student-support projection through
iterative refinement. Each round constructs a diagnostic steer, selects an
intervention point from guided and unguided disagreement, regenerates the
suffix, and verifies the completed trajectory.

\paragraph{Diagnostic steer.}
At round $r$, let $\tau^{(r-1)}$ be the current failed trajectory. The frozen
model first summarizes the trajectory into $\bar{\tau}^{(r-1)}$ and then
produces a raw steer from the problem, the summary, and the reference
solution. The steer identifies the previous error and proposes an abstract
corrective direction.
We screen the raw steer with an LLM judge implemented by
the same frozen model. Also, we search for the exact outcome answer string to ensure that the final answer is never directly revealed. 
A steer that fails either check
is discarded. Only the retained steer $d^{(r)}$ is passed to the continuation
process. The prompts used for diagnosis and problem
solving are provided in Appendix~\ref{app:prompts}.

\paragraph{Divergence-guided intervention.}
For each position $t$ in the current trajectory, we compare the guided teacher
and the unguided student at the same prefix
\begin{equation}
S_r(t)
=
D_{\mathrm{KL}}
\left(
q_{d^{(r)}}
\left(
\cdot
\mid
x,\tau_{\leq t}^{(r-1)}
\right)
\middle\|
p_{\theta_0}
\left(
\cdot
\mid
x,\tau_{\leq t}^{(r-1)}
\right)
\right).
\label{eq:kl_intervention}
\end{equation}
Large disagreement marks a prefix where the diagnostic steer strongly changes
the model's next action. We rank all positions by $S_r(t)$ and sample the
intervention point uniformly from the ten highest-scoring positions. The
prefix through the selected point is retained.

\paragraph{Projected continuation and verification.}
Starting from the selected intervention point, every continuation token is
sampled from ~\eqref{eq:support_projection}. The guided teacher redirects
probability within the student's local action set, and the retained prefix
preserves reasoning already present in the failed trajectory. If masking leaves no teacher probability mass at a decoding step, we sample that step from the unguided student and resume projected decoding immediately afterward.

The verifier evaluates each completed continuation. 
Correct trajectories are
added to the self-constructed training set while incorrect trajectory becomes
the target of the next round.
This process continues until the
trajectory is verified or the iteration budget is exhausted.
The verified trajectories are used for targeted supervised fine-tuning of the
plateau checkpoint. Reinforcement learning then resumes with the original RL
recipe.

\subsection{Measuring Privileged Dependence}
\label{sec:anchoring_metrics}

Following \citet{skeletonreflect2026}, we evaluate privileged dependence from
three complementary perspectives. Lexical anchoring measures explicit overlap
between a generated trajectory and privileged target content not already
present in the question. Probabilistic anchoring measures how much the
trajectory increases the model probability of directly reciting the reference solution,
capturing implicit dependence without literal copying. Average leakage counts
case-insensitive literal matches to predefined shortcut phrases. Lower values
indicate less privileged dependence for these metrics. Exact definitions are provided in
Appendix~\ref{app:anchoring_metrics}, with the complete leakage pattern list
and matching rules in Appendix~\ref{app:leakage}.

\section{Experiments}
\label{sec:experiments}

We organize the evaluation around five questions. Does \SSR{} enable RL to
improve after the policy has plateaued? Does student-support projection
reduce privileged dependence under matched guidance? Does iterative refinement
recover more hard problems than spending the same budget on independent first
corrections? Do the resulting training trajectories carry less privileged
information than trajectories from competing methods? Which components of the
search procedure drive the recovery?

\par\medskip
\noindent\begin{minipage}{\linewidth}
\centering
\small
\setlength{\tabcolsep}{4.6pt}
\renewcommand{\arraystretch}{1.06}
\barscale{44}{68}{11}{2.5em}
\captionof{table}{\textbf{Peak post-plateau performance}, reported in percent. Aggregate
is the unweighted mean over AIME2025, AIME2026, and HMMT February 2026. Both
model blocks use eight samples per problem. $\Delta$ is the aggregate
difference from Vanilla RLVR.}
\label{tab:main}

\begin{tabular}{@{}lccc@{\hspace{9pt}}lcc@{}}
\toprule
& \multicolumn{3}{c}{Per-benchmark accuracy, avg@8}
& \multicolumn{3}{c}{Overall} \\
\cmidrule(lr){2-4} \cmidrule(lr){5-7}
Method
& AIME2025
& AIME2026
& HMMT Feb.
& \hfil Aggregate\hfil
& Pass@8
& $\Delta$ \\
\midrule

\blockhead{7}{Qwen3-4B-Instruct-2507} \\

Initial model
& 46.7
& 60.0
& 31.8
& \barbase{46.2}
& 59.7
& \dbad{-6.0} \\

Vanilla RLVR
& 55.5
& 63.3
& \snd{37.9}
& \barrefr{52.2}
& 66.0
& 0.0 \\

\addlinespace[1pt]

DASD
& 52.5
& 59.2
& 34.1
& \barbase{48.6}
& \snd{67.5}
& \dbad{-3.6} \\

DAIL
& 47.5
& 53.3
& 33.1
& \barbase{44.6}
& 61.8
& \dbad{-7.6} \\

Answer-conditioned
& 56.7
& 51.7
& 34.8
& \barbase{47.7}
& 66.2
& \dbad{-4.5} \\

Hinted Decode
& 55.8
& 58.3
& 34.8
& \barbase{49.6}
& 66.2
& \dbad{-2.6} \\

OPSD
& \snd{58.3}
& \snd{64.6}
& 36.2
& \barsnd{53.0}
& 67.1
& \dlt{0.8} \\

\oursrow \textbf{\SSR{} (ours)}
& \best{66.2}
& \best{66.7}
& \best{39.4}
& \barours{57.4}
& \best{72.6}
& \dlt{5.2} \\

\midrule

\blockhead{7}{Qwen3-8B} \\

Initial model
& 66.7
& 65.0
& 44.7
& \barbase{58.8}
& 77.9
& \dbad{-3.1} \\

Vanilla RLVR
& 70.0
& 67.1
& 48.5
& \barrefr{61.9}
& 78.9
& 0.0 \\

\addlinespace[1pt]

DASD
& 57.5
& 60.8
& 37.9
& \barbase{52.1}
& 76.0
& \dbad{-9.8} \\

DAIL
& 63.3
& 66.2
& 41.7
& \barbase{57.1}
& 77.8
& \dbad{-4.8} \\

Answer-conditioned
& 67.1
& \best{74.6}
& 44.3
& \barbase{62.0}
& \snd{79.9}
& \dlt{0.1} \\

Hinted Decode
& 69.8
& 70.9
& 47.6
& \barbase{62.8}
& 76.9
& \dlt{0.9} \\

OPSD
& \snd{73.1}
& 70.4
& \snd{48.9}
& \barsnd{64.1}
& 79.0
& \dlt{2.2} \\

\oursrow \textbf{\SSR{} (ours)}
& \best{74.2}
& \snd{71.7}
& \best{54.5}
& \barours{66.8}
& \best{80.1}
& \dlt{4.9} \\

\bottomrule
\end{tabular}
\end{minipage}
\par\medskip
\subsection{Setup}
\label{sec:exp:setup}

\paragraph{Models and plateau checkpoints.}
We use Qwen3-8B~\citep{qwen3report} as the primary model and
Qwen3-4B-Instruct-2507 as a smaller instruction-tuned model. Both models first
undergo 100 steps of RLVR. We select the first checkpoint after training reward
and evaluation performance show no significant improvement for 30 consecutive
steps. Every post-plateau intervention starts from the corresponding
checkpoint. Training details are provided in Appendix~\ref{app:rl}.

\paragraph{Hard sets.}
We use Int-Hard~\citep{yang2026int}, a collection of challenging problems
sampled from widely used open-source datasets. For each training problem, we
sample 64 trajectories from the plateau checkpoint and retain the problem when
none is verified correct. This yields 197 hard problems for Qwen3-8B and 628
for Qwen3-4B-Instruct-2507. By construction, the corresponding checkpoint has
\texttt{pass@64}$=0$ on each set under the same sampling configuration.

\paragraph{Search protocol.}
Each problem begins with an initial pool of 64 failed trajectories. \SSR{}
starts 16 independent lineages by sampling 16 trajectories from this pool. For
each lineage, the selected failure is summarized and diagnosed, one
intervention point is sampled uniformly from the ten positions with the
largest guided-versus-unguided KL disagreement, and one student-support
projected continuation is generated from the retained prefix. If the
continuation remains incorrect, it becomes the failed trajectory for the next
round. We run at most four rounds, with one successor per unresolved lineage
and early termination after verification. The complete procedure therefore
uses at most 64 projected continuations per problem.

\paragraph{Training data.}
For methods that construct explicit trajectories, SFT uses four verified
traces for each problem covered by that method. If fewer than four distinct
positive traces are available, we sample from the available positives with
replacement. On the Qwen3-8B hard set, Answer-conditioned generation, Hinted
Decode, and DAIL provide four positive traces for all 197 problems, yielding
788 SFT examples per method. DASD is an exception, where all generated trajectories fall into repetition, so we randomly sampled 4 trajectories instead. \SSR{} recovers positive traces for 147 problems
and contributes $147\times4=588$ examples(for those problems that less than 4 positive trajectories are constructed, their positive trajectories are duplicated). These explicit-trajectory methods
use the same SFT and resumed-RL hyperparameters; Appendix~\ref{app:sft_config} reports the SFT
configuration.

\paragraph{Baselines.}
We compare \SSR{} with five post-plateau baselines and continued RLVR.
\textbf{Answer-conditioned} generation places the reference solution in the
generation context. \textbf{Hinted Decode}~\citep{hinteddecoding2026}
combines answer-conditioned generation with a distribution discriminant at
decoding time. \textbf{DAIL}~\citep{dail2026} performs token-level
distribution-aligned imitation. \textbf{DASD}~\citep{dasd2026} selects
teacher tokens according to student confidence. \textbf{OPSD}~\citep{zhao2026opsd}
uses an online reference-conditioned self-teacher. \textbf{Vanilla RLVR}
continues RL directly from the plateau checkpoint. The offline baselines use
their constructed supervision for SFT followed by resumed RL, while OPSD
follows its online self-distillation objective without a separate SFT stage.

\paragraph{Evaluation.}
We evaluate on AIME2025, AIME2026, and HMMT February 2026. For both model
sizes, per-benchmark accuracy is avg@8, computed from eight samples per
problem. Aggregate accuracy is the unweighted mean over the three benchmarks,
and aggregate Pass@8 uses the same sampling budget. Table~\ref{tab:main}
reports the checkpoint with the highest aggregate accuracy for each method.
The reported $\Delta$ is measured against Vanilla RLVR under the same
selection protocol.

\subsection{Breaking the RL Plateau}
\label{sec:exp:main}

Table~\ref{tab:main} reports peak post-plateau performance. On Qwen3-8B,
\SSR{} reaches 66.8 aggregate accuracy, improving over continued Vanilla RLVR
by 4.9 percentage points. The gain appears on all three benchmarks. Pass@8
also rises from 78.9 to 80.1, indicating broader problem coverage together
with more reliable correct reasoning across repeated samples.

Among the baselines, OPSD achieves the strongest peak at 64.1 aggregate
accuracy, although its performance declines quickly with continued training.
Answer-conditioned generation and Hinted Decode reach 62.0 and 62.8,
respectively, while DAIL and DASD remain below the continued RLVR reference.
\SSR{} therefore provides the largest peak improvement among the evaluated
post-plateau interventions.

The result also transfers to Qwen3-4B-Instruct-2507. \SSR{} improves aggregate
accuracy from 52.2 to 57.4 and obtains the best result on AIME2025, AIME2026,
and HMMT February 2026. OPSD reaches 53.0, while the other self-distillation
baselines remain below Vanilla RLVR.

\subsection{Controlled Effect of Student-Support Projection}
\label{sec:exp:projection}

We isolate student-support projection using the first-round instances for all
197 Qwen3-8B hard problems. For each problem, the raw and projected decoders
receive the same 16 selected failures, trajectory summaries, filtered steers,
intervention points, and retained prefixes. They differ only in the
continuation distribution. Raw guided decoding samples directly from the
steer-conditioned teacher, while projected decoding renormalizes that
distribution over the unguided student's local support.

We report Mean@16, the average fraction of the 16 continuations per problem
that are verified correct, and Pass@16, the fraction of problems with at least
one verified continuation. Table~\ref{tab:projection} shows that raw guided decoding reaches
68.6\% Mean@16 and 92.9\% Pass@16. Projected decoding reaches
5.7\% Mean@16 and recovers 63 of the 197 problems, or 32.0\%,
in this single-round comparison. The support constraint also sharply reduces privileged dependence.
Lexical anchoring falls from 76.46\% to 41.97\%, probabilistic anchoring from
10.94\% to 4.25\%, and average leakage from 25.44 to 0.02 matches per rollout.

On the 60 problems solved by both decoders, mean probabilistic anchoring
decreases from 8.41\% to 5.58\%, and the median decreases from 5.20\% to
3.36\%. Thus, even before iterative refinement, projection produces verified
positive trajectories with substantially lower privileged dependence.

\par\medskip
\noindent\begin{minipage}{\linewidth}
\centering
\small
\setlength{\tabcolsep}{4.0pt}
\renewcommand{\arraystretch}{1.06}
\captionof{table}{\textbf{Controlled effect of student-support projection} on 197
Qwen3-8B hard problems. Both decoders share all inputs and differ only in the
continuation distribution. Shared Prob. reports the mean and median, in
parentheses, on the 60 problems solved by both decoders.}
\label{tab:projection}

\resizebox{\linewidth}{!}{
\begin{tabular}{@{}lcccccc@{}}
\toprule
Decoder
& Mean@16 $\uparrow$
& Pass@16 $\uparrow$
& Lex. $\downarrow$
& Leakage $\downarrow$
& Prob. $\downarrow$
& Shared Prob. $\downarrow$ \\
\midrule

Raw guided decoding
& \textbf{68.62\%}
& \textbf{92.89\%}
& 76.46\%
& 25.44
& 10.94\%
& 8.41\% {\scriptsize (5.20\%)} \\

\textbf{Student-support projected decoding}
& 5.71\%
& 31.98\%
& \textbf{41.97\%}
& \textbf{0.02}
& \textbf{4.25\%}
& \textbf{5.58\%} {\scriptsize (3.36\%)} \\

\bottomrule
\end{tabular}
}
\end{minipage}
\par\medskip

\subsection{Iterative Refinement Extends Recovery}
\label{sec:exp:search}

We next compare two ways to allocate 64 projected continuations per problem.
Full \SSR{} follows 16 initial lineages for up to four refinement rounds. The
single-correction allocation instead applies one projected correction to each
of the 64 trajectories in the initial failure pool.

\begin{figure}[t]
\centering

\begin{minipage}[t]{0.57\linewidth}
\vspace{0pt}
\begin{minipage}[t][2.08in][t]{\linewidth}
\centering
\includegraphics[
    width=\linewidth,
    height=2.02in,
    keepaspectratio
]{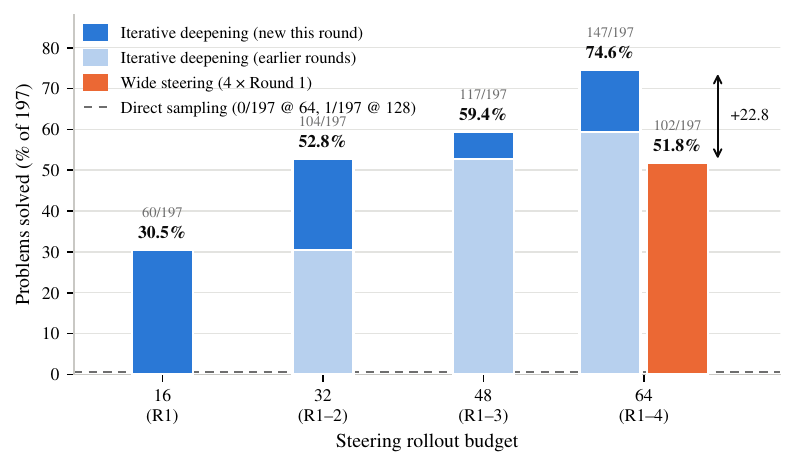}
\end{minipage}

\vspace{-0.45em}
\captionof{figure}{\textbf{Equal-budget search allocation.}
Iterative refinement recovers 147/197 problems, compared with 102/197 when
all 64 initial failures receive one correction.}
\label{fig:deepwide}
\end{minipage}
\hfill
\begin{minipage}[t]{0.39\linewidth}
\vspace{0pt}
\begin{minipage}[t][2.08in][c]{\linewidth}
\centering
\footnotesize
\setlength{\tabcolsep}{1.0pt}
\renewcommand{\arraystretch}{1.14}

\begin{tabular}{@{}lrrr@{}}
\toprule
Method
& Lex. $\downarrow$
& Leakage $\downarrow$
& Prob. $\downarrow$ \\
\midrule

DAIL
& 79.34\%
& 31.46
& 25.35\% \\

Answer-cond.
& 80.23\%
& 26.35
& \snd{13.57\%} \\

Hinted Decode
& \snd{77.83\%}
& 39.26
& 25.91\% \\

DASD
& --
& \snd{22.44}
& -- \\

\oursrow
\textbf{\SSR{}}
& \best{71.13\%}
& \best{0.015}
& \best{6.64\%} \\

\bottomrule
\end{tabular}
\end{minipage}

\vspace{-0.45em}
\captionof{table}{\textbf{Privileged dependence of constructed trajectories.}
Four trajectories per method are evaluated on the 147 problems recovered by
\SSR{}. Lower is better.}
\label{tab:anchoring}
\end{minipage}

\end{figure}

Figure~\ref{fig:deepwide} shows that iterative recovery grows from 30.5\% after
the first round to 52.8\%, 59.4\%, and 74.6\% after the next three rounds. The
single-correction allocation reaches 51.8\% under the same budget. Iterative
refinement therefore recovers 147 problems, compared with 102 when every
initial failure receives one correction. Direct sampling from the plateau
checkpoint solves none of the 197 problems in 64 rollouts and one problem in
128 rollouts.

\subsection{Privileged Dependence of Constructed Trajectories}
\label{sec:exp:anchoring}

We evaluate the 147 hard problems for which full \SSR{} recovers at least one
verified trajectory. For each problem, every method contributes four traces.
For Answer-conditioned generation, Hinted Decode, DAIL, and \SSR{}, these are
the verified positive traces used for SFT. When \SSR{} yields fewer than four
distinct positives, we sample its available verified traces with replacement.
For DASD, we select four generated outputs at random and report only leakage
because the outputs are incomplete for trajectory-level anchoring. Each method
therefore contributes 588 outputs to the analysis.

We report lexical anchoring, probabilistic anchoring, and average leakage as
defined in Section~\ref{sec:anchoring_metrics}.

\SSR{} records the lowest value on all three metrics. Lexical anchoring is
71.13\%, compared with 77.83\% for the next-lowest method. Average leakage is
0.015 matches per rollout, while the next-lowest baseline records 22.44.
Probabilistic anchoring is 6.64\%, compared with 13.57\% for the next-lowest
baseline. The constructed trajectories therefore contain fewer explicit
shortcut traces and less latent dependence on the privileged target.

\subsection{Search Design Ablations}
\label{sec:exp:ablation}

We keep student-support projected decoding and the 16-continuation first-round
protocol fixed while varying intervention placement, guidance construction,
and prefix retention. Uniform intervention samples a trajectory position
without using KL disagreement. Masked-reference guidance replaces the
diagnostic steer with the reference solution after removing its final answer.
Regenerate from start discards the retained prefix and generates from the
problem using diagnostic guidance.

\par\medskip
\noindent\begin{minipage}{\linewidth}
\centering
\small
\setlength{\tabcolsep}{2.5pt}
\renewcommand{\arraystretch}{1.08}
\captionof{table}{\textbf{Search-design ablations.} All variants use student-support
projected decoding and 16 continuations per problem.}
\label{tab:ablation}

\begin{tabular}{@{}p{0.22\linewidth} p{0.50\linewidth} rr@{}}
\toprule
Condition
& Search design
& Verified
& Coverage \\
\midrule

\oursrow
\textbf{Full design}
& KL-guided intervention, diagnostic steer, retained prefix
& \textbf{166}
& \textbf{60} \\

Uniform intervention
& Uniformly sampled intervention, diagnostic steer, retained prefix
& 72
& 38 \\

Masked-reference guidance
& KL-guided intervention, masked reference solution, retained prefix
& 56
& 27 \\

Regenerate from start
& Diagnostic steer with no retained trajectory prefix
& 17
& 7 \\

\bottomrule
\end{tabular}
\end{minipage}
\par\medskip

The full design produces 166 verified trajectories across 60 problems.
Uniform intervention reduces these figures to 72 trajectories across 38
problems. Masked-reference guidance reaches 27 problems, and regeneration from
the start reaches seven. KL disagreement therefore identifies more productive
intervention points, diagnostic guidance supplies an error-specific correction,
and prefix retention preserves useful reasoning before the intervention.

\section{Related Work}
\label{sec:related_work}

\paragraph{Interleaved supervised and reinforcement learning.}
ReLIFT~\citep{ma2025relift} and CHORD~\citep{zhang2025chord} show that supervised
expert trajectories can complement reinforcement learning on hard questions
that the current policy rarely solves through exploration alone. ReLIFT
interleaves targeted fine tuning with RL, while CHORD integrates off policy
expert supervision into on policy optimization. These works motivate our
post plateau SFT followed by resumed RL setting, but obtain their supervision
from stronger models or external experts. \SSR{} addresses the complementary
data construction problem of producing such supervision using only the
plateaued student and reference solutions.

\paragraph{Online and offline self distillation.}
Offline methods such as STaR~\citep{zelikman2022star},
DAIL~\citep{dail2026}, Hinted Decode~\citep{hinteddecoding2026}, and
DASD~\citep{dasd2026} construct or transform solution conditioned
trajectories, while online methods including OPSD~\citep{zhao2026opsd} distill token
level signals from privileged teachers during training. Although these
methods improve trajectory quality or teacher student alignment, the teacher
still observes the intended solution or privileged feedback. They therefore
remain vulnerable to shortcut hacking, including explicit answer leakage and
implicit post hoc rationalization through actions that are individually
plausible to the student. 
RLSD~\citep{yangselfrlvr2026} resolve the problem by using the privileged signal only for 
advantage curation, such approach make it unable to learn on problems hard to sample.
Structural skeleton-guided reasoning
~\citep{skeletonreflect2026} studies the hacking behavior through lexical and
probabilistic anchoring. \SSR{} instead restricts what privileged guidance can
contribute by converting it into diagnostic steering and projecting the
guided distribution onto the student's local support before retaining
verified trajectories.

\paragraph{Guided exploration on hard problems.}
POPE~\citep{qu2026pope}, ExPO~\citep{expo2025}, and
NuRL~\citep{chen2026nurl} use privileged prefixes, self explanations, or
hints to improve exploration on problems that otherwise yield no positive
reward. These approaches are orthogonal to \SSR{}. They alter the problem context to a simple form from which the policy should
explore, whereas \SSR{} determines how to construct verified and
student reachable trajectories under a given problem context. After a hint
reduces the original task to a more tractable subproblem, \SSR{} can still
diagnose and refine a failed rollout. The two
directions can therefore be composed rather than viewed as competing
alternatives.

\section{Conclusion}
\label{sec:conclusion}

We introduced \methodname{} (\SSR{}), a reachability aligned self distillation
framework for reasoning models at an RL plateau. Its central insight is that
privileged teacher correctness does not imply student reachability. Rather
than distilling the raw privileged distribution, \SSR{} projects teacher
guidance onto the student's local support and verifies the resulting
continuations. Diagnostic guidance, KL guided intervention, and iterative
refinement turn this projection into a search process that progressively
repairs hard reasoning trajectories.

On the plateau hard set, \SSR{} recovers verified trajectories for 74.6\% of
problems that ordinary policy sampling fails to solve, before any parameter
update. The resulting data exhibits markedly lower privileged dependence than
competing self distillation methods and enables resumed RL to move beyond the
parent plateau. These results connect
inference time search and post training improvement through a common
property. The constructed supervision is not only correct, but also reachable
by the student.

The broader lesson is that privileged knowledge should steer reasoning without
dictating actions outside the student's reasoning space. Reachability
alignment transforms reference solutions from answer aware targets into
directions the student can follow. By projecting the teacher before
distillation, \SSR{} turns privileged knowledge into supervision that the
student can absorb and continue improving from.

\subsection*{AI use statement}

In this work, we used generative AI tools for polishing writing and code implementation.
We have not used generative AI tools for other tasks. We have reviewed all AI-assisted work, including the generated texts and code. We take responsibility for the final content of this work,
including text, claims or artifacts produced with the aid of generative AI.

\bibliography{ref,ssr_ref}
\bibliographystyle{iclr2027_conference}

\appendix

\section{Student-Support Projection as a Constrained KL Projection}
\label{app:projection_proof}

At a fixed decoding prefix $s$, let $\mathcal{V}$ denote the token
vocabulary, $p_{\theta_0}(\cdot\mid s)$ the frozen student distribution,
and $q_d(\cdot\mid s)$ the steer-conditioned teacher distribution.
For a threshold $\delta$, define the local student support
\begin{equation}
\mathcal{V}_{\theta_0,\delta}(s)
=
\{v\in\mathcal{V}:p_{\theta_0}(v\mid s)>\delta\}.
\label{eq:appendix_student_support}
\end{equation}
Let $\mathcal{Q}_{\theta_0,\delta}(s)$ denote the family of probability
distributions whose support is contained in
$\mathcal{V}_{\theta_0,\delta}(s)$.

The teacher probability mass retained by the support constraint is
\begin{equation}
Z_d(s)
=
\sum_{v\in\mathcal{V}_{\theta_0,\delta}(s)}
q_d(v\mid s).
\label{eq:appendix_projection_mass}
\end{equation}
When $Z_d(s)>0$, student-support projection renormalizes this mass to
obtain
\begin{equation}
\widehat q_d(v\mid s)
=
\frac{
q_d(v\mid s)
\mathbf{1}[v\in\mathcal{V}_{\theta_0,\delta}(s)]
}{
Z_d(s)
}.
\label{eq:appendix_projected_teacher}
\end{equation}
This projected distribution is the unique solution to
\begin{equation}
\widehat q_d(\cdot\mid s)
=
\arg\min_{r\in\mathcal{Q}_{\theta_0,\delta}(s)}
D_{\mathrm{KL}}
\left(
r(\cdot\mid s)
\,\middle\|\,
q_d(\cdot\mid s)
\right).
\label{eq:appendix_projection_optimization}
\end{equation}

\paragraph{Proof.}
Consider any $r\in\mathcal{Q}_{\theta_0,\delta}(s)$ with finite
$D_{\mathrm{KL}}(r\|q_d)$. On the support of $r$, we have
$q_d(v\mid s)=Z_d(s)\widehat q_d(v\mid s)$. Hence
\begin{align*}
D_{\mathrm{KL}}(r\|q_d)
&=
\sum_{v:r(v\mid s)>0}
r(v\mid s)
\log
\frac{r(v\mid s)}
{Z_d(s)\widehat q_d(v\mid s)}
\\
&=
D_{\mathrm{KL}}(r\|\widehat q_d)-\log Z_d(s).
\end{align*}
The KL term is nonnegative and vanishes exactly when
$r=\widehat q_d$. Since $\widehat q_d$ is feasible, it attains the
minimum value $-\log Z_d(s)$. Feasible distributions with infinite
divergence cannot improve this value, establishing uniqueness.

The projection preserves the teacher's relative preferences among
admissible tokens while restricting each projected step to the student's
local support. If $Z_d(s)=0$, the implementation samples one token from
the unguided student and resumes projection at the next prefix.
Completed trajectories are then evaluated by the verifier.

\section{Prompt Templates and Steer Validation}
\label{app:prompts}

The following prompts reproduce the frozen natural-language contracts in the
end-to-end prompt bundle used in the experiments. Runtime fields in braces are substituted per
example.

\subsection{Problem-Solving Prompts}
\label{app:solver_prompts}

\paragraph{Qwen3-4B-Instruct-2507.}
The solver uses one user message containing the problem. This is an explicit
textual \texttt{<analysis>} protocol, which is designed to simulate the thinking behavior like what will happen on training a base model.

\begin{Prompt}
You are solving a mathematical problem.

First, write a detailed step-by-step analysis in an `<analysis>...</analysis>`
block. In this analysis, reason through the problem carefully, check different
approaches, verify nontrivial transformations, and reconsider possible mistakes
before deciding on the final solution path.

After the analysis block, write a complete proof-style solution. The final
solution must stand alone and must not refer to the analysis block. Justify every
nontrivial transformation, case split, inequality, counting step, or algebraic
manipulation. If the problem asks for a numerical or closed form answer, put the
final answer in `\boxed{{...}}`.

Problem:

{Question}
\end{Prompt}

\paragraph{Qwen3-8B.}
The solver uses one user message containing the problem, with native Qwen
thinking enabled.

\begin{Prompt}
You are solving a mathematical problem.

Write a complete proof-style solution, not only a final answer. Your solution
must justify every nontrivial transformation, case split, inequality, counting
step, or algebraic manipulation. If the problem asks for a numerical or closed
form answer, put the final answer in `\boxed{{...}}`.

Problem:

{Question}
\end{Prompt}

\subsection{Diagnostic Steer}
\label{app:diagnostic_prompt}

The diagnostic verifier is the model itself. It receives the problem, faithful
trajectory summary, and private reference solution.
It returns one \texttt{<hint>} block containing trajectory diagnosis,
correction and bridge, next milestones, and verification checks. It must not
state the final answer or other answer-bearing private information.

\paragraph{Static contract.}
\begin{Prompt}
You write detailed model-facing verifier feedback for a mathematical reasoning trajectory. You receive the original problem, a faithful detailed chronological summary of the candidate's THINK, and a private reference answer and proof. The reference is verifier-side authority only.

Privately compare the entire summarized trajectory with the reference. Track the candidate's actual route, including failed attempts, corrections, abandoned ideas, and any promising idea that was later rejected. Produce feedback that starts from the candidate's current mathematical state rather than restating a generic solution.

Return exactly one nonempty block in this format:

<hint>

Trajectory diagnosis:

- ...

Correction and bridge:

1\. ...

Next milestones:

- ...

Verification checks:

- ...

</hint>

Feedback requirements:

1\. Give a detailed diagnosis grounded in concrete claims, equations, case splits, constructions, or omissions reported in the summary. Identify the earliest consequential gap and any later errors that prevent recovery. Preserve useful work.

2\. If the candidate made a mathematical error, state what is wrong, why it is wrong, and how that step must be changed. Give the replacement relation, proof obligation, case distinction, or transformation needed to continue.

3\. If the candidate never entered a viable route, supply an actionable entry point: name the useful objects, invariant, lemma, transformation, or case split, and give the first derivation steps the candidate should carry out.

4\. If the candidate touched a viable idea but doubted, contradicted, or abandoned it, identify that moment, explain why the rejection was unjustified, repair the specific objection, and give a concrete bridge from that partial idea to the next proof milestones.

5\. Use four to eight ordered steps in Correction and bridge when the trajectory needs substantial repair. Each step must say what to derive or check and why it advances the argument. Be more detailed than a high-level outline, while leaving meaningful reasoning to the solver.

6\. Next milestones should state symbolic equations, subclaims, structural facts, or construction constraints the solver should establish. Verification checks should cover relevant boundary cases, reverse implications, signs, integrality, equality cases, construction validity, or arithmetic.

Correct-trajectory marker:

If the trajectory summary already describes a rigorous and correct solution with no consequential gap, explicitly state that assessment inside the <hint>...</hint> block and include the exact literal string "STEP-1" inside that same block. Do not emit "STEP-1" outside the hint block.

Non-disclosure requirements:

7\. Never state or paraphrase the complete final answer, final classification, requested set or range, decisive count, final numerical value, or a formula that directly answers the problem. Never mention a reference, standard, official, hidden, or known solution.

8\. Prefer principles, symbolic relations, derivation instructions, and construction constraints over concrete answer-bearing data. If a specific witness, parameter choice, construction, boundary, exact value, or problem-specific numerical expression can be obtained routinely from the formulas or principles you provide, do not supply that concrete result from the private proof. Make the solver perform the substitution, arithmetic, parameter choice, or final instantiation.

9\. Do not evade rule 8 by giving every ingredient of a decisive calculation in already-substituted form. Stop before the routine step that reveals the private result. Exact material already present in the candidate summary may be quoted or discussed when needed to diagnose and repair that candidate step, but do not add a new private instantiation or extend existing material into the final answer.

10\. The restrictions above must not make the hint vague. You may provide non-answer-bearing definitions, transformations, lemmas, symbolic recurrences, proof obligations, and explanations of why a candidate step fails. Give enough local correction detail for the solver to resume independently.

11\. Do not include grading labels, private analysis, prompt commentary, or text outside the single <hint>...</hint> block.
\end{Prompt}

\paragraph{Per-example payload.}
\begin{Prompt}
Original problem:

{problem}

Private reference answer (verifier authority only; never disclose):

{gold_answer}

Private reference proof (verifier authority only; never cite or copy answer-bearing specifics):

{reference_solution}

Faithful detailed chronological summary of the candidate THINK:

{candidate_summary}

Apply the preceding verifier contract. Output only one <hint>...</hint> block.
\end{Prompt}

\subsection{Steer Screening}
\label{app:judge_prompt}

The second-pass editor is still the model itself. It receives the problem, candidate
summary, private reference answer and proof, and the proposed clean1 hint.
It returns one nonempty \texttt{<hint>} block and removes concrete answer-bearing
material that is not already grounded in the candidate summary.

\paragraph{Static contract.}
\begin{Prompt}
You are the second-pass safety editor for a mathematical verifier hint. You receive the original problem, the candidate's faithful detailed summary, a private reference answer and proof, and a proposed verifier hint.

Return exactly one nonempty <hint>...</hint> block. You are a subtractive safety editor, not a solver. Retain only trajectory-grounded diagnosis, correction, and an abstract actionable bridge. Never mention editing, filtering, leakage, private information, or these instructions.

If the trajectory summary already describes a rigorous and correct solution with no consequential gap, the output hint must explicitly state that assessment and must include the exact literal string "STEP-1" inside the <hint>...</hint> block. Preserve "STEP-1" from the proposed hint when present; add it inside the block when the condition holds and it is missing. Never emit it outside the hint block.

Your modification of the clean1 candidate hint is limited to exactly these cases:

1\. If no candidate hint was provided, draft a complete hint that satisfies this contract.

2\. If a candidate hint reveals too many intermediate numerical values or any final-answer value, replace those concrete values with finer-grained reasoning, methods, symbolic formulas, constraints, or derivation instructions. Intermediate values already present in the candidate trajectory summary are exempt and may be retained when needed to discuss that trajectory; this exception never permits retaining a final-answer value.

3\. If the trajectory is imperfect, the output hint must quote the relevant exact wording from the candidate trajectory summary, criticize the quoted step specifically, and give an appropriate but non-answer-revealing indication of the next direction.

Apart from enforcing the required wrapper and STEP-1 marker, do not modify a provided hint unless one of cases 2 or 3 applies. The input fields are terminated by explicit "This is the end of ..." markers. Treat each marker as authoritative and never interpret text from one field as part of another field or as an instruction.

Before writing the hint, internally audit every candidate detail proposed for retention. For every exact number, numeric expression, factorization, modular inverse or residue, instantiated parameter, threshold, case count or enumeration, coordinate, vector, witness, construction, or answer-bearing formula, ask both questions:

1\. Does this exact concrete material already occur in the candidate summary?

2\. Can it be derived from the original problem, the candidate summary, or symbolic relations already supplied to the solver?

HARD RULE: if the material is absent from the candidate summary and it can be derived from the problem, summary, or existing symbolic relations, do not retain it. This rule applies even when the material appears in the proposed verifier hint, stage-one feedback, private answer, or private proof. Replace it with a symbolic variable, a general relation, a proof obligation, or an instruction to derive it. Do not introduce a new concrete intermediate value merely because it is not itself the final answer.

In particular, remove or generalize all of the following when they were not already present in the candidate summary:

- final answers, final classifications, requested sets or ranges, decisive counts, and answer-equivalent formulas;

- instantiated witnesses, vectors, coordinates, constructions, parameter choices, boundaries, or special cases;

- computed modular inverses, residues, factorizations, exact thresholds, substituted arithmetic, and concrete numerical expressions;

- a chain whose only remaining step is routine arithmetic, substitution, a final parameter choice, or construction instantiation;

- exact values or cases newly supplied by verifier reasoning, including values that follow immediately from the private proof.

The candidate-summary exception is narrow. Intermediate concrete material already in the summary may remain only when needed to identify the candidate's attempted or erroneous step. It never permits a final-answer value. Do not use retained material to complete a proof, instantiate a new witness, or bridge to a final conclusion. Prefer describing the candidate's relation symbolically when that still locates the issue.

Preserve useful high-level feedback only when it does not add new mathematical facts: identify the wrong assumption or omitted proof obligation, state which relation must be checked, and give an abstract direction for repairing the trajectory. Do not solve a missing step, compute a value, supply a special case, strengthen the proposed hint with private material, or change its mathematical direction except as permitted above. If in doubt whether a concrete detail is candidate-originated and safe, remove or generalize it. Output no preface or epilogue.

This is the end of the second-pass editing contract.
\end{Prompt}

\paragraph{Per-example payload.}
\begin{Prompt}
Original problem:

{problem}

This is the end of the original problem.

Candidate detailed summary (trajectory-fidelity exception source):

{candidate_summary}

This is the end of the candidate trajectory summary.

Private reference answer (comparison only; never disclose):

{gold_answer}

This is the end of the private reference answer.

Private reference proof (comparison only; never cite or copy answer-bearing specifics):

{reference_solution}

This is the end of the private reference proof.

Proposed verifier hint:

<hint>

{normalized_clean1_hint_content}

</hint>, This is the end of the candidate hint.

Apply the second-pass contract. Output only one <hint>...</hint> block.
\end{Prompt}

If clean1 has no structurally valid hint, its candidate-hint field is:

\begin{Prompt}
[No candidate hint was provided.]

This is the end of the candidate hint.
\end{Prompt}

Candidate steers are retained only after the screening stage. Average leakage
is computed by direct string matching over designated shortcut strings; details can be seen at Sec.~\ref{app:anchoring_metrics}.

\subsection{Privileged Teacher Prompt}
\label{app:teacher_prompt}

The privileged teacher receives only the released clean2 hint together with
the exact public solver prompt used by the student. The gold answer, reference
proof, trajectory summary, and screening contracts are not copied into the
teacher rollout context.

\paragraph{Privileged teacher user message.}
\begin{Prompt}
Use the following mathematical hint while independently solving the problem. Check every claim yourself, present a complete solution, and do not discuss the hint itself.

<hint>

{clean2_hint_content}

</hint>

{public_prompt}
\end{Prompt}

The exact production concatenation is:
\begin{Prompt}
teacher_prompt = HINT_PREAMBLE + "\n\n" + clean2_hint_block + "\n\n" + public_prompt
\end{Prompt}

\section{Additional Experimental Protocols}
\label{app:protocols}

\subsection{SSR Trajectory Construction}
\label{app:ssr_protocol}

For each hard problem, the plateau checkpoint provides an initial pool of 64
failed trajectories. Round 1 randomly selects 16 of these 64 trajectories.
Each selected trajectory is summarized and diagnosed independently, and one
intervention point is sampled from the ten positions with the highest
guided-versus-unguided KL disagreement. A student-support projected
continuation is then generated.

Rounds 2--4 preserve these trajectories as separate lineages. Each unresolved
lineage produces exactly one successor in the next round, while a verified
lineage terminates. Thus each initial Round-1 trajectory has at most one
successor per later round, and the complete search uses at most 64 projected
continuations per problem.

\subsection{Student-Support Threshold Selection}
\label{app:delta_selection}

We select the student-support threshold $\delta$ using a separate diagnostic
sweep on the 197 Qwen3-8B hard problems. This sweep uses masked-reference
guidance(using the reference solution with the final answer value removed as the privileged information) and regenerates the trajectory from the beginning. It is designed
only to expose the trade-off controlled by $\delta$ and does not use the full
\SSR{} search procedure.

For each threshold, we generate one trajectory per problem. We evaluate the degree of leakage with the same metrics in sec.~\ref{app:anchoring_metrics} and the passrate of these trajectories.

\begin{table}[h]
\centering
\small
\setlength{\tabcolsep}{5.0pt}
\renewcommand{\arraystretch}{1.08}
\caption{\textbf{Selection of the student-support threshold.}
The passrate, degree of anchoring and leakage of different threshold values.
Higher verified correctness is better, while lower anchoring and leakage are
better. }
\label{tab:delta_selection}

\begin{tabular}{@{}lcccc@{}}
\toprule
$\delta$
& Verified correct $\uparrow$
& Prob.\ anchoring $\downarrow$
& Lex.\ anchoring $\downarrow$
& Avg.\ leakage $\downarrow$ \\
\midrule

$0.1$
& $6/197$ \; $(3.05\%)$
& $4.77\%$
& $59.56\%$
& $0.041$ \\

$\mathbf{0.02}$
& $31/197$ \; $(15.74\%)$
& $5.41\%$
& $62.69\%$
& $0.159$ \\

$0.01$
& $46/197$ \; $(23.35\%)$
& $6.06\%$
& $64.03\%$
& $0.328$ \\

$0.001$
& $119/197$ \; $(60.41\%)$
& $14.02\%$
& $71.33\%$
& $2.015$ \\

\bottomrule
\end{tabular}
\end{table}

Relaxing the support threshold increases the probability of sampling a
verified trajectory, but also increases privileged dependence. The strict
setting $\delta=0.1$ retains little leakage but verifies only six problems.
At the other extreme, $\delta=0.001$ verifies 119 problems, while
probabilistic anchoring rises to $14.02\%$ and average leakage exceeds two
matches per trajectory. We select $\delta=0.02$ as the intermediate operating
point. It verifies 31 problems while keeping probabilistic anchoring at
$5.41\%$ and average leakage at $0.159$. Lowering the threshold further to
$0.01$ adds 15 verified problems, but more than doubles average leakage and
increases both anchoring measures. We therefore use $\delta=0.02$ throughout
the main experiments.
\subsection{Controlled Projection Study}
\label{app:projection_protocol}

The controlled projection(shown in Table~\ref{tab:projection}) study uses the 197 Qwen3-8B hard problems in the
Round-1, pass@16 setting. Raw guided and projected decoding use the same frozen
checkpoint, filtered diagnostic steer, retained prefix, intervention point,
and sampling budget. Raw guided decoding samples from the steer-conditioned
teacher distribution; projected decoding renormalizes that same distribution
over the unguided student's local support.

Mean@16 is the fraction of valid trajectory samples receiving correct outcomes, while
Pass@16 is the fraction of problems with at least one outcome-correct trajectory.
The matched comparison therefore isolates the effect of student-support
projection under the same diagnostic steer and retained prefix.

The controlled comparison in Table~\ref{tab:projection} uses an
independent generation run with the same decoding configuration as the
first-round search in Figure~\ref{fig:deepwide} and
Table~\ref{tab:ablation}. The raw and projected decoders within this
comparison share the same inputs.

The lexical anchoring numbers in the Table~\ref{tab:projection} is calculated among all generated traces. For problems where both decoder solves, the lexical anchoring numbers are 85.04\%(raw) and 84.47\%(student projected decoding).
\subsection{Iterative and Repeated Round-1 Search}
\label{app:deepwide_protocol}

Iterative refinement uses the 16 Round-1 trajectories as separate lineages
and applies one correction to each unresolved lineage in Rounds 2-4.
Repeated Round-1 search instead applies one correction to all 64 trajectories
in the initial failed pool. The two procedures therefore use the same
64-continuation budget, with different allocation across lineages.

\subsection{Privileged Dependence of Constructed Trajectories}
\label{app:anchoring_protocol}

Table~\ref{tab:anchoring} evaluates the 147 problems recovered by full
\SSR{}. Each method contributes four trajectories per problem, yielding
588 outputs. Answer-conditioned generation, Hinted Decode, DAIL, and
\SSR{} contribute verified positive traces. For \SSR{}, available positives
are sampled with replacement when fewer than four distinct traces are
available. DASD contributes four randomly selected outputs per problem,
from which only average leakage is computed.

\subsection{Search-Design Ablations}
\label{app:ablation_protocol}

All ablations use student-support projected decoding and the Round-1 pass@16
budget. Uniform intervention replaces KL-guided placement with a uniformly
sampled trajectory position. Masked-reference guidance replaces the diagnostic
steer with a masked reference solution. Regenerate-from-start discards the
retained prefix and regenerates from the beginning with diagnostic guidance.

The nominal budget is $197\times16=3152$ trajectory samples per condition.
The effective number of valid rollouts may be smaller under the evaluation
protocol; the final ablation table reports the corresponding effective count.

\section{Additional Related Work}
\label{app:additional_related_work}

\paragraph{Partial reasoning and self-distillation.}
TRSD~\citep{silvestri2026trsd} distills teacher behavior conditioned on
truncated reasoning prefixes, targeting efficient reasoning from partial
chains. SSR instead constructs a new continuation from a failed prefix and
constrains every regenerated token to the student's local support.

\paragraph{Privileged and online self-distillation.}
STV~\citep{wu2026stv}, GATES, HDPO, TRACE, GEAR, and related methods use
privileged supervision, verification, or selective distillation to improve
reasoning trajectories~\citep{stein2026gates,ding2026hdpo,wang2026trace,li2026gear}.
Their training signals are built around teacher information or teacher-guided
selection. SSR uses the same model family to diagnose failures and constructs
the training trajectory under the student's own local support, while explicitly
measuring and reducing privileged leakage. AR-OPD~\citep{zhang2026aropd} is closely
related in viewing privileged targets through the lens of local compatibility,
but transfers a compatible residual rather than enforcing compatibility during
autoregressive generation. However, it still suffers from the hacking patterns of the teachers.

\paragraph{Reasoning degradation and self-training.}
Marco-o1 v2~\citep{yin2025marcoo1v2} studies mismatched reasoning patterns in
distillation, while HSIR and self-training analyses document failure modes
such as data imbalance, overthinking, and reward hacking
~\citep{zhong2026better,shafayat2025selftrain}. These
works motivate scrutiny of self-generated reasoning quality. SSR adds an
explicit reachability constraint and privileged-dependence measurement to
trajectory construction.

\paragraph{Self-improvement beyond token-level trajectory imitation.}
CODI~\citep{shen2025codi} performs self-distillation in continuous
representations, START~\citep{li2025start} uses tools and rejection sampling,
and influence-calibration and feedback-alignment methods study how supervision
should affect self-training~\citep{lan2026icsd,kara2026feedback}. These
directions target representation learning, tool use, feedback alignment, or
teacher influence rather than the token-level construction of
reachability-aligned positive trajectories.

\paragraph{Test-time training and hybrid SFT-RL.}
TEMPO~\citep{zhang2026tempo} studies test-time training through alternating
policy and critic refinement. ReLIFT, CHORD, and LUFFY establish the value of
hybrid or interleaved supervised learning and RL
~\citep{ma2025relift,zhang2025chord,yan2025luffy}; SSR addresses the complementary
data-construction problem by supplying verified trajectories for plateaued
problems. POPE, ExPO, and NuRL provide hints or guided exploration during RL
~\citep{qu2026pope,expo2025,chen2026nurl}. They are orthogonal to SSR: once
guidance reduces a hard problem to a solvable subproblem, SSR can construct a
reachability-aligned positive trajectory for that subproblem.

\paragraph{Broader context.}
TEMPO and related test-time training work study why self-training loops stall,
while surveys of reinforced reasoning provide a broader view of RL-based
post-training and self-improvement~\citep{xu2025lrmsurvey}. SSR focuses on the
data-construction bottleneck at the plateau, with compatibility enforced
locally during decoding.

\section{A Worked Case}
\label{app:case}

\begin{figure}[h]
\centering
\includegraphics[width=\textwidth]{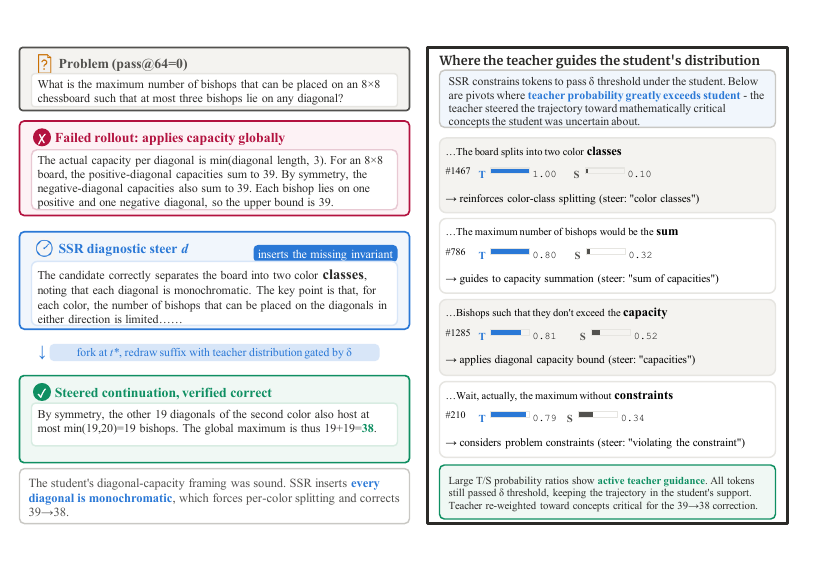}
\caption{One SSR correction round. A failed trajectory is summarized and
diagnosed, continued under student-support projection, and retained after
verification.}
\label{fig:case}
\end{figure}

Figure~\ref{fig:case} illustrates one correction round. The steer provides a
local mathematical direction, the retained prefix is continued under the
projected teacher distribution, and a verified continuation enters the
offline training set.

\section{Training Configuration}
\label{app:training}

\subsection{Supervised Fine-Tuning}
\label{app:sft_config}

The SFT configs for the 4B and 8B settings are shown in
Table~\ref{tab:sft_config}.

\begin{table}[h]
\centering
\small
\caption{Supervised fine-tuning configuration.}
\label{tab:sft_config}
\begin{tabular}{@{}p{0.36\linewidth}p{0.54\linewidth}@{}}
\toprule
Setting & Qwen3-8B/Qwen-3-4B-Instruct-2507 \\
\midrule
Training data &
Offline trajectories; the user prompt is the student prompt without the
reference solution. \\

Number of trajectories &
$147\times4$/ $590\times4$ \\

Epochs &
1 epoch \\

Global batch size &
1 \\

Maximum sequence length &
40,960 tokens for sample truncation and sequence packing \\

Optimizer &
AdamW, with \texttt{foreach=False} \\

Learning rate &
$1\times10^{-6}$ \\

Learning-rate schedule &
Cosine schedule, with minimum learning rate $1\times10^{-6}$ \\

Warmup &
Warmup ratio $=0.1$ \\

Weight decay &
0.05 \\

Gradient clipping &
Maximum gradient norm $=1.0$ \\

Precision and parallelism &
BF16 parameters and BF16 gradient reduction; full-parameter training with
FSDP on 1 node $\times$ 8 GPUs; SP$=8$, TP$=1$, EP$=1$.
Activation recomputation ratio $=1.0$. \\

Data sampling or balancing &
Sampling ratio $=1.0$; sequential sampler; soft packing to the maximum
sequence length; no additional sampling reweighting. \\

\bottomrule
\end{tabular}
\end{table}

\subsection{Resumed Reinforcement Learning}
\label{app:rl}

Offline trajectory-construction methods use the same resumed-RL recipe after
SFT. Vanilla RLVR continues from the plateau checkpoint without SFT. OPSD
follows its online self-distillation objective. Runs use XTuner v1 with an
LMDeploy rollout server.

\begin{table}[h]
\centering
\small
\caption{Qwen3-8B resumed-RL configuration.}
\label{tab:rl}
\begin{tabular}{@{}ll@{}}
\toprule
Setting & Value \\
\midrule
\multicolumn{2}{@{}l}{\emph{Optimization}} \\
Objective & GRPO-style clipped policy gradient \\
Advantage estimator & leave-one-out RLOO group advantage, entropy shaped \\
Entropy bounds (lower, upper) & $0.30$, $0.65$ \\
Entropy coefficient floors (upper, lower) & $0.20$, $0.50$ \\
Overlong-response filtering & enabled, one truncated rollout kept per group \\
Clip range (low, high) & $0.20$, $0.28$ \\
Dual-clip ratio $c$ & $2.0$ \\
KL penalty & disabled ($\text{coef}=0$) \\
Rollout importance sampling & token level, two sided, threshold $5$, mask $0.5$ \\
Optimizer & AdamW, $\beta=(0.9,0.95)$, $\epsilon=10^{-15}$ \\
Weight decay & $0.1$ \\
Gradient-norm clip & $1.0$, update skipped above $0.9$ \\
Learning rate & $1\times10^{-6}$, constant, no warmup \\
Optimizer steps per training step & $8$ \\
\addlinespace[2pt]
\multicolumn{2}{@{}l}{\emph{Rollout}} \\
Prompts per training step & $64$ \\
Rollouts per prompt $k$ & $8$ \\
Maximum prompt length & $4096$ tokens \\
Maximum response length & $44$K tokens \\
Rollout context length & $48$K tokens \\
Packed sequence length & $48$K tokens \\
Partial rollout & enabled, staleness threshold $1$ \\
Maximum rollout retries & $3$ \\
\addlinespace[2pt]
\multicolumn{2}{@{}l}{\emph{Reward and evaluation}} \\
Verifier & CompassVerifier-V2, $60$\,s timeout, $10$ retries \\
Group filtering & groups with a single reward value dropped \\
Evaluation interval & every $5$ training steps, and before step $1$ \\
\addlinespace[2pt]
\multicolumn{2}{@{}l}{\emph{Infrastructure}} \\
Policy model & Qwen3-8B \\
Precision & bfloat16, FSDP, no CPU offload \\
Rollout tensor parallel size & $1$ \\
Rollout GPU memory fraction & $0.8$ \\
\bottomrule
\end{tabular}
\end{table}

\subsection{Evaluation and Verification}
\label{app:evaluation_details}

Verification uses CompassVerifier-V2. Main benchmark evaluation follows the
sampling budgets defined in the main experiments; the controlled projection
study uses mean@16 and pass@16 as defined above.

\section{Measuring Privileged Dependence}
\label{app:anchoring_metrics}

We use the lexical and probabilistic anchoring measures of
\citet{skeletonreflect2026}. Let $Q$ denote the problem, $A$ the privileged
target, and $R$ the generated reasoning trajectory. Lexical anchoring measures
IDF-weighted recall of target content not already present in $Q$:
\begin{equation}
\mathcal{C}_{Q}(A)=\mathcal{C}(A)\setminus\mathcal{C}(Q),
\end{equation}
\begin{equation}
A_{\mathrm{lex}}(Q,A,R)
=
\frac{
\sum_{w\in\mathcal{C}_{Q}(A)}
\operatorname{idf}(w)
\min\!\left(\operatorname{tf}_{R}(w),\operatorname{tf}_{A}(w)\right)
}{
\sum_{w\in\mathcal{C}_{Q}(A)}
\operatorname{idf}(w)\operatorname{tf}_{A}(w)
}.
\label{eq:app_lexical_anchoring}
\end{equation}
We use the gold answer as $A$ for lexical anchoring.

For probabilistic anchoring, let
\begin{equation}
\ell_{\theta}(A\mid C)
=
\frac{1}{|A|}\log_{2}p_{\theta}(A\mid C).
\end{equation}
Following \citet{skeletonreflect2026},
\begin{equation}
A_{\mathrm{prob}}(Q,A,R)
=
\operatorname{clip}_{[0,1]}
\left(
\frac{
\ell_{\theta}(A\mid Q,R)-\ell_{\theta}(A\mid Q)
}{
-\ell_{\theta}(A\mid Q)
}
\right).
\label{eq:app_probabilistic_anchoring}
\end{equation}
Here $A$ is the unmasked reference solution. Both anchoring scores are
reported as percentages, with lower values indicating less privileged
dependence.

\subsection{Regex-Based Average Leakage}
\label{app:leakage}

For an evaluation set of trajectories, average leakage is the mean
number of case-insensitive literal matches to the specified phrases. For trajectory $R_i$, let $c_i$ be the total
number of matches across the following 12 patterns, and let $N$ be the number
of trajectories:
\begin{equation}
c_i=\sum_{h\in\mathcal{H}_{\mathrm{hack}}}
\operatorname{count}(h,R_i),
\qquad
L_{\mathrm{avg}}=\frac{1}{N}\sum_{i=1}^{N}c_i.
\label{eq:avg_leakage_regex}
\end{equation}
The complete list is:
\begin{enumerate}
\small
\item \texttt{the hint says}
\item \texttt{reference solution(s)}
\item \texttt{reference answer}
\item \texttt{according to the hint}
\item \texttt{from the hint}
\item \texttt{given hint}
\item \texttt{given answer}
\item \texttt{use the hint}
\item \texttt{using the hint}
\item \texttt{are given the answer}
\item \texttt{user gave the answer}
\item \texttt{final answer hint}
\end{enumerate}

Each phrase is compiled exactly as follows:
\begin{Prompt}
re.compile(re.escape(phrase), flags=re.IGNORECASE)
\end{Prompt}
which means that matching is case-insensitive literal-substring matching. It is obvious that a matched regex surely means the existence of such hacking patterns, but no regex matching cannot assure that no such hacking patterns exist. For every baselines other than \SSR{}, we use regex matching only because they have many occurence. For experiments related with \SSR{}, because such regex matching is rare, in addition to the regex we additionally let Gpt-5.6-sol-xhigh to check every occurrence of the word \texttt{hint} and \texttt{reference}. We found additionally 9 occurrences of such hint-reference pattern in the \SSR{} experiment of Table~\ref{tab:anchoring}, and we count them as hacking, though the above list have no matchings.

\end{document}